\documentclass{article}
\usepackage{ijcai22}

\usepackage{times}
\usepackage{soul}
\usepackage{url}
\usepackage[hidelinks]{hyperref}
\usepackage[utf8]{inputenc}
\usepackage[small]{caption}
\usepackage{graphicx}
\usepackage{amsmath}
\usepackage{amsthm}
\usepackage{booktabs}
\usepackage{algorithm}
\usepackage{algorithmic}
\usepackage{amsfonts}
\usepackage{multirow}
\usepackage{subfigure}
\usepackage{subfloat}
\title{Efficient and Accurate Conversion of Spiking Neural Network with Burst Spikes}

\iftrue
\author{
Yang Li$^{1,2}$
\And
Yi Zeng$^{1,2,3,4}$\footnote{Corresponding author}
\affiliations
$^1$Research Center for Brain-inspired Intelligence, Institute of Automation, Chinese Academy of Sciences\\
$^2$School of Artificial Intelligence, University of Chinese Academy of Sciences\\
$^3$Center for Excellence in Brain Science and Intelligence Technology, Chinese Academy of Sciences\\
$^4$National Laboratory of Pattern Recognition, Institute of Automation, Chinese Academy of Sciences\\
\emails
\{liyang2019, yi.zeng\}@ia.ac.cn
}
\fi

\begin{document}

\maketitle

\begin{abstract}
     Spiking neural network (SNN), as a brain-inspired energy-efficient neural network, has attracted the interest of researchers. While the training of spiking neural networks is still an open problem. One effective way is to map the weight of trained ANN to SNN to achieve high reasoning ability. However, the converted spiking neural network often suffers from performance degradation and a considerable time delay. To speed up the inference process and obtain higher accuracy, we theoretically analyze the errors in the conversion process from three perspectives: the differences between IF and ReLU, time dimension, and pooling operation. We propose a neuron model for releasing burst spikes, a cheap but highly efficient method to solve residual information. In addition, Lateral Inhibition Pooling (LIPooling) is proposed to solve the inaccuracy problem caused by MaxPooling in the conversion process. Experimental results on CIFAR and ImageNet demonstrate that our algorithm is efficient and accurate. For example, our method can ensure nearly lossless conversion of SNN and only use about 1/10 (less than 100) simulation time under 0.693$\times$ energy consumption of the typical method. Our code is available at \href{https://github.com/Brain-Inspired-Cognitive-Engine/Conversion_Burst}{\textit{https://github.com/Brain-Inspired-Cognitive-Engine/Conversion\_Burst}}.
    
\end{abstract}

\section{Introduction}

As a representative of artificial intelligence, deep learning has surpassed human performance in many fields. However, the artificial neural network (ANN) training procedure requires enormous energy consumption and massive memory, which is challenging to apply to lightweight devices and limited storage scenarios. Recently, neuromorphic hardware such as TrueNorth \cite{akopyan2015truenorth} and Loihi \cite{davies2018loihi} has attracted the attention of researchers due to their low computational burden, and fast processing \cite{roy2019towards}. This kind of hardware does not run artificial neural networks but spiking neural networks (SNNs) \cite{maass1997networks}, which is considered the third generation neural network. It uses discrete spikes for information transmission and inference based on the spatio-temporal paradigm, so the calculation is more efficient and biologically plausible.

Although the spiking neural network has the above advantages, the backpropagation algorithm widely used in ANN cannot be used for directly training the spiking neural networks due to the non-differentiable character of the spikes. So far, training highly efficient spiking neural networks remains an open problem. Some progress has been made in training networks directly using synaptic plasticity rules \cite{zeng2017improving}, or surrogate gradient methods \cite{wu2018spatio}. The former draws on the brain's nervous system rules to make spiking neural networks more explicable. The efficient selection and combination of these learning rules can help us explore the nature of intelligence \cite{zeng2017improving}. The latter uses mature backpropagation technology to smooth the spike propagation process. The gradient optimization-based SNN effectively combines their advantages, which helps explore effective spatio-temporal feature representation \cite{wu2018spatio}. However, these methods show poor scalability on large networks and datasets. The training process needs enormous computing resources and simulation time.

\begin{figure}[!t]
	\centering
	\includegraphics[scale=0.25]{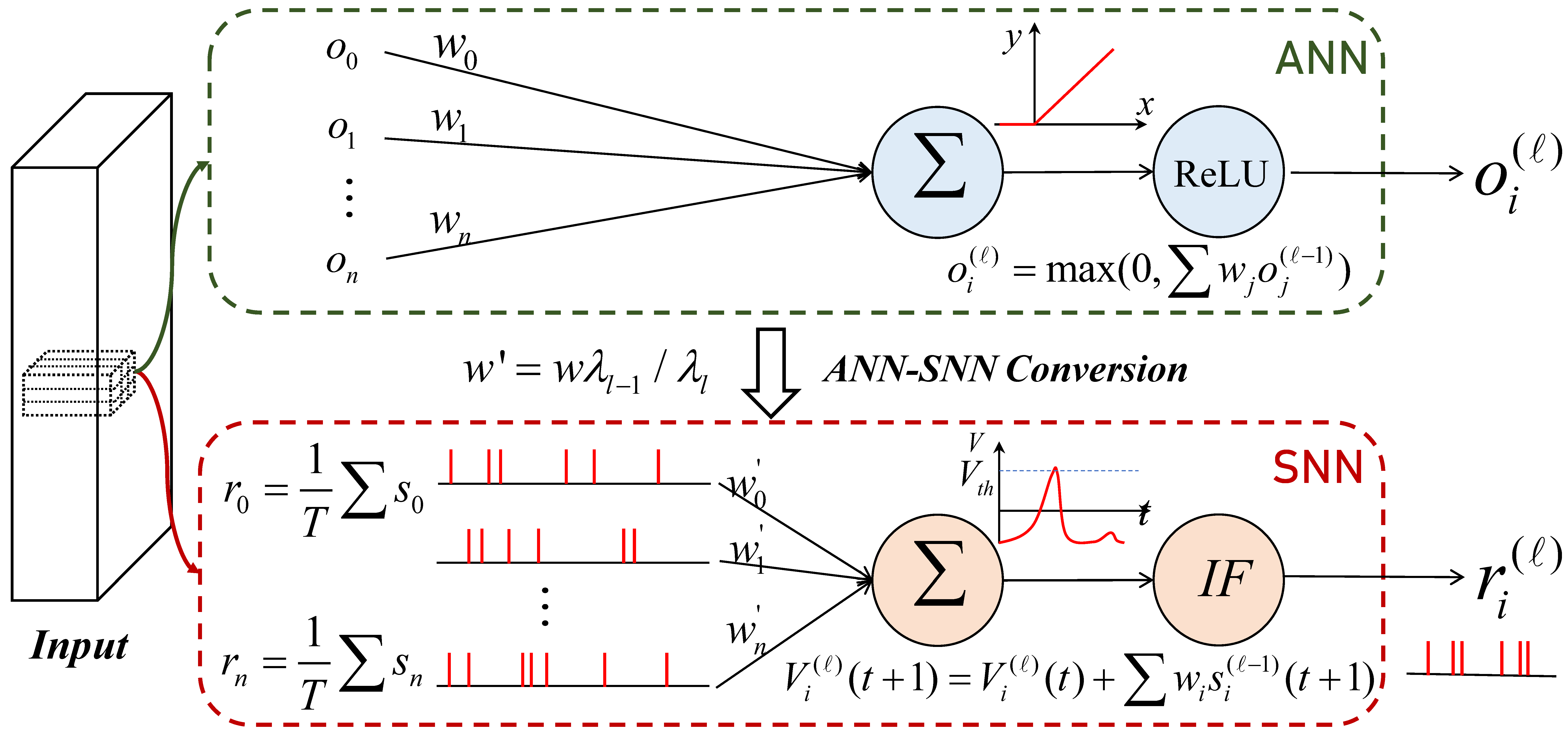}
	\caption{Illustration of the ANN-SNN conversion.}
	\label{fig1}
\end{figure}

To obtain efficient inference ability and demonstrate the low energy consumption characteristics of SNN, the researchers convert well-trained ANN to SNN \cite{cao2015spiking,diehl2015fast}, which significantly reduces the training time and calculation consumption. Fig. \ref{fig1} shows the ANN-SNN conversion process. So far,  the converted SNN with Integrate-and-Fire neurons (IF) has demonstrated great advantages in reinforcement learning, object detection, and tracking \cite{tan2020strategy,kim2020spiking,luo2020siamsnn}. However, the conversion process often suffers from performance losses and time delays. Methods such as soft-reset, weight normalization, and spike-norm \cite{han2020rmp,diehl2015fast,sengupta2019going} have been proposed to address these problems but still require long simulation times (usually greater than 256 or even thousands) to achieve acceptable performance. It greatly limits the use of SNN on mobile devices and in specific scenarios, such as mobile phones and target tracking. As far as we know, it is unclear how the converted weights affect the performance of SNN. Except for the neuron differences between ANN and SNN, the temporal dimension of SNN and other operations such as the pooling layer receive little attention.

In order to obtain excellent performance in dozens of simulation times, we theoretically analyze the error of the conversion process from the perspective of SNN structure and propose a conversion strategy using burst spikes to make the firing rate in SNN approximate to the activation value in ANN. Meanwhile, Lateral Inhibition Pooling (LIPooling) based on lateral inhibition mechanism \cite{koyama2018mutual} is proposed to replace MaxPooling. To our best knowledge, this is the first time to analyze the error from the whole conversion process, including differences between ReLU and IF, time dimension, and pooling opration perspectives. Moreover, the above method does not need additional layers in the ANN training process. Our main contributions can be summarized as follows:

\begin{itemize}
	\item We formulate the conversion process and divide the conversion errors into residual information, spikes of inactivated neurons, and pooling errors. Then, we analyze their influence on the conversion process.
	\item We propose a neuron model for releasing burst spikes under the clock-driven simulation framework and propose the LIPooling to make information transfer in SNN more accurate.
	\item The experimental results demonstrate the accuracy and high efficiency of the proposed algorithm. Our method achieves state-of-the-art performance in less than 100 time steps.
\end{itemize}

\section{Related Work}
	The synaptic plasticity and surrogate gradient methods often require more computing resources and are ineffective. While the ANN-SNN conversion method proposed by Cao et al. \shortcite{cao2015spiking} can adapt to more complex situations. Diehl et al. \shortcite{diehl2015fast} attribute the conversion error to inaccurate activation values of neurons and propose data-based weight normalization and threshold balancing methods. Spike-Norm \cite{sengupta2019going} also tries threshold balancing from the SNN perspective. In order to balance the simulation time and accuracy and make the conversion process more robust, Rueckauer et al. \shortcite{rueckauer2017conversion} propose the p-Norm method. They also extend bias and Batch Normalization (BN) to the conversion process. The soft-reset method is proposed to further reduce the information loss in the generation of spikes. RMP method \cite{han2020rmp}with soft reset achieves near-lossless conversion through threshold balancing. Spiking-YOLO \cite{kim2020spiking} uses channel-wise weight normalization to achieve object detection but requires more than 2k simulation steps. Zhang et al.     \shortcite{zhang2019tdsnn} and Han et al. \shortcite{han2020deep} try to use temporal coding for conversion. They only use one spike or two spikes to further reduce the network's energy consumption. Kim et al. \shortcite{kim2018deep} and Li et al.\shortcite{li2021bsnn} try to use phase coding to pack more information to the spikes but still need a long simulation time. The hybrid method \cite{rathi2020enabling} achieves good performance by using the synaptic plasticity or surrogate gradient method to finetune the converted SNN. Ding et al. \shortcite{ding2021optimal} make the output of ReLU more convenient for IF approximation by adding a limiting layer in the training process of ANN, but the problems in the conversion process are still not solved. As far as we know, it is still tough to get high-performance SNN with low time delay. In addition, most of the work tends to use average pooling without addressing inaccuracies of MaxPooling in the conversion process.

\section{Method}

In this section, we formulate the conversion process. Based on the analysis of conversion error, we propose the neuron model with burst spikes and LIPooling to solve the problems of residual information and pooling error, respectively. Then, experiments are carried out to prove the efficiency and accuracy of the proposed method. 

\begin{figure*}[!t]
	\centering
	\includegraphics[scale=0.5]{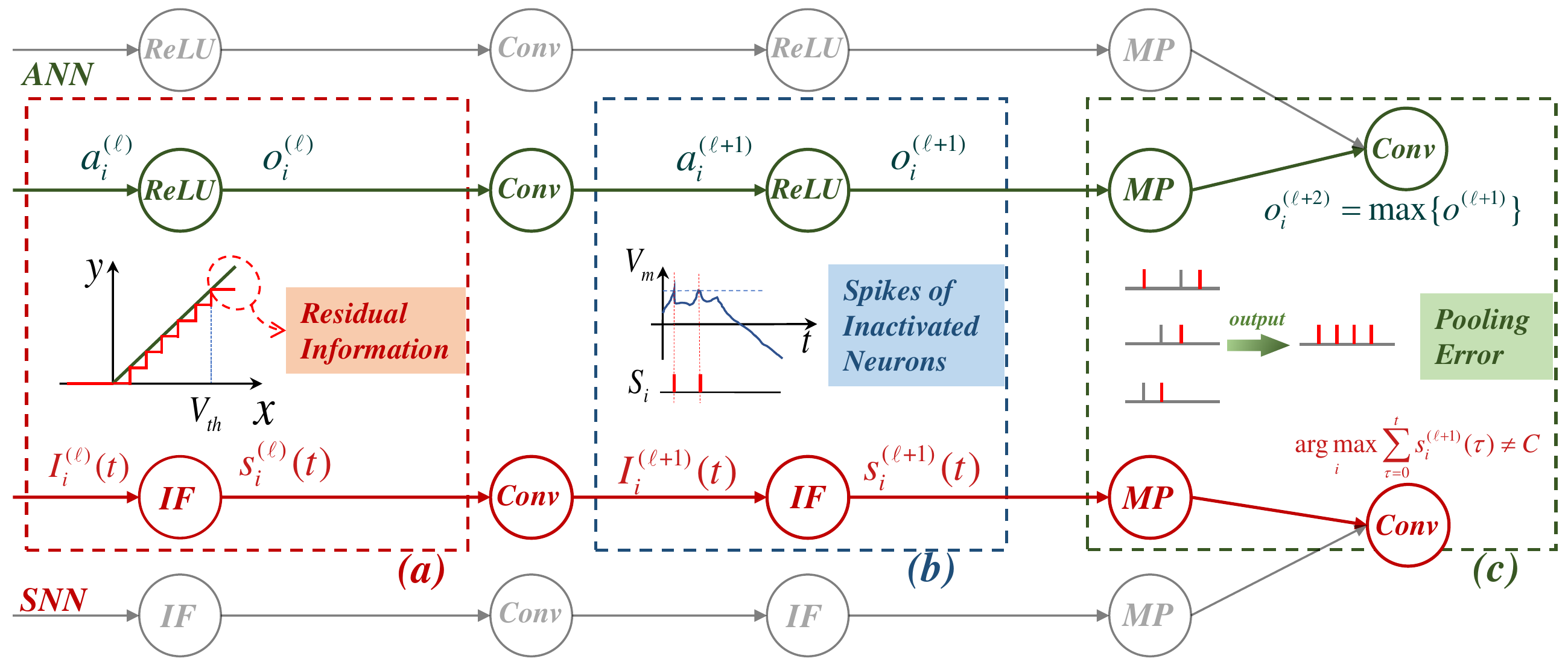}
	\caption{The conversion error from IF neuron, time dimension, and MaxPooling. (a) The maximum firing rate of IF neurons is one; the remaining membrane potential cannot be released by spikes. And the accuracy is limited by simulation time. (b) In SNN, due to the instability of synaptic current, neurons whose activation value is 0 will also emit spikes when they reach the threshold in a short time. (c) SNN uses spikes of the neuron with the maximum firing rate to do MaxPooling, but the neuron is changing. It cannot correct the historical error firing, resulting in a larger output than the actual value.}
	\label{error}
\end{figure*}

\subsection{Conversion from ANN to SNN}

\paragraph{Neuron Model for ANN.} For the $\ell$-th convolutional layer or fully connected layer in ANN, its forward propagation process can be described as follows:
\begin{align}
	o^{(\ell)}_i = g(a^{(\ell)}_i) = g(\textbf{w}_i^{(\ell)}\textbf{o}^{(\ell-1)}), \quad 1 \leq \ell \leq L
\end{align}
where $\textbf{a}^{(\ell)}$ and $\textbf{o}^{(\ell)}$ represent the input and the activation value in $\ell$-th layer respectively. $g(x)$ is the ReLU activation function. $\textbf{w}^{(\ell)}$ denotes the weight, and bias $\textbf{b}^{(\ell)}$ is omitted from the pre-activation values for ease of representation. $L$ is the number of layers of the network.
\paragraph{Neuron Model for SNN.} We use Integrate-and-Fire (IF) neuron for conversion. At the time step $t$, the IF neuron integrates the binary spike information of the previous layer. $\textbf{w}_i^{(\ell)}\textbf{s}^{(\ell-1)}$ can be regarded as the synaptic current $I_i^{(\ell)}$ of the neuron $i$. The membrane potential changes after receiving the current as follows:
\begin{align}
	V_i^{(\ell)}(t) = 	V_i^{(\ell)}(t-1) + \textbf{w}_i^{(\ell)}\textbf{s}^{(\ell-1)}
\end{align}
where $V_i^{(\ell)}$ means the membrane potential at time step $t$. When the neuron membrane potential exceeds the threshold $V_{th}^{(\ell)}$, which is usually set to 1, the neuron fires and its membrane potential is updated by soft-reset \cite{rueckauer2017conversion}, i.e., the threshold is subtracted. Conversely, the membrane potential remains constant, and $s_i^{(\ell)}(t)=0$.
\begin{align}
	s_i^{(\ell)}(t) = &
	\begin{cases}
		1 \quad if \quad V_i^{(\ell)}(t) \geq V_{th}^{(\ell)},\\
		0 \quad else.
		\label{spike}
	\end{cases}
\end{align}
\begin{align}
	V_i^{(\ell)}(t) = V_i^{(\ell)}(t) - s_i^{(\ell)}(t)
	\label{vm}
\end{align}
\paragraph{ANN-SNN Conversion.} The principle of conversion is to use the firing rate in SNN to approximate the activation value in ANN. Ideally, the two are approximately equal after T time steps, as shown below.
\begin{align}
	\label{principle}
	o_i^{(\ell)} \approx r_i^{(\ell)} = \frac{1}{T}\sum \limits_{t=0}^T s_i^{(\ell)}(t) 
\end{align}
The maximum firing rate in SNN is one under the time-driven simulation framework. It requires that the activation values in ANN be distributed between 0 and 1. In order to solve this problem, data-based weight normalization \cite{diehl2015fast} is proposed. We use the $p$-th percentile of the activation values in the $\ell$-th layer $\lambda_{\ell}=\max_p^{(\ell)}$ to scale the corresponding weight. Since the bias is additive, there is no need to multiply the maximum value of the previous layer.
\begin{align}
	\label{norm}
	\hat{\textbf{w}}^{(\ell)} = \textbf{w}^{(\ell)}
	\frac{\lambda_{\ell-1}}{\lambda_{\ell}},\quad \hat{\textbf{b}}^{(\ell)} = \frac{\textbf{b}^{(\ell)}}{\lambda_{\ell}}
\end{align}
As for batch normalization, we follow \cite{rueckauer2017conversion} and merge the convolutional layer and the subsequent BN layer to form a new convolutional layer. 
\begin{align}
	\label{equ7}
	\hat{\textbf{w}}^{(\ell)}=\frac{\gamma_i}{\theta_i}\textbf{w}^{(\ell)}, \quad \hat{\textbf{b}}^{(\ell)}=\frac{\gamma_i}{\theta_i}(\textbf{b}^{(\ell)}- \mu_i)+\beta_i.
\end{align}
An input $\textbf{a}$ is transformed into $BN[\textbf{a}]=\frac{\gamma}{\theta}(\textbf{a}-\mu)+\beta$, where $\mu$ and  $\theta$ are mean and variance of batch, $\beta$ and $\gamma$ are two learned parameters during training.

\subsection{Dividing the Conversion Loss}

\label{32}

Consider the case where weights are normalized using the parameter $p$, then the neuron in $\ell$-th can be divided into $\mathbb{R}_1= \left\{j \mid \frac{o_{j}^{(\ell)}}{max_{p}^{(\ell)}} \leq 1\right\}$ which represents the neuron set whose activation value of layer $\ell$-th is less than 1 after data normalization and $\mathbb{R}_2=\left\{j \mid \frac{o_{j}^{(\ell)}}{max_{p}^{(\ell)}} > 1\right\}$ whose activation value is more than 1. The total received synaptic current in $\ell$-th layer at time step $t$ can be written as
\begin{equation}
	\begin{aligned}
		\sum_{\tau=0}^{t} I_{i}^{(\ell)}(\tau)
		=&\sum_{\tau=0}^{t} \sum \hat{w}_{i j} s_{j}^{(\ell-1)}(\tau)\\
		=&\sum_{j \in \mathbb{R}_1} \frac{w_{i j}}{max_{p}^{(\ell)}}  \sum_{\tau=0}^{t} s_{j}^{(\ell-1)}(\tau)+\sum_{j \in \mathbb{R}_2} \frac{w_{i j}}{max_{p}^{(\ell)}}  t
		\label{residual}
	\end{aligned}
\end{equation}

It can be seen from Eq. (\ref{residual})  that clip operation for activation values greater than 1 causes \textbf{residual information} in the membrane potential of neurons at the previous layer, affecting the accuracy of information transmission, as shown in the red box in Fig. \ref{error}. In addition, the choice of $p$ will make the conversion suffer from the dilemma of accuracy and speed. When $p$ is larger, there are more neurons in $\mathbb{R}_1$ set, but less information of current $I$ can be integrated at each time step.

Previous works do not take the influence of time dimension on conversion into account, and consider $\sum\limits_{t=0}^Ts_i^{(\ell)}(t)=\left\lfloor\frac{\sum_{t=0}^T I^{(\ell)}_i(t)}{V_{th}}\right\rfloor $ when $ \frac{o_{j}^{(\ell)}}{max_{p}^{(\ell)}} \in \{0,1\}$, where $\lfloor x \rfloor$ returns the largest integer smaller than $x$ . Ideally, we hope $I_i^{(\ell)}(t)=a_i^{(\ell)}$, then Eq. (\ref{principle}) will follow. However, as the synapse current changes from the weighted sum to the integration of discrete spikes in SNN, $\sum\limits_{\tau=0}^t I_i^{(\ell)}(\tau)$ fluctuates around $t \cdot a_i^{(\ell)}$ with time $t$. When $\sum\limits_{\tau=0}^t I_i^{(\ell)}(\tau)\geq V_{th}^{(\ell)}$, neurons with $o_i^{(\ell)}=0$ can also emit spikes, called \textbf{spikes of inactivated neurons (SIN) }as shown in the blue box in Fig. \ref{error}. Suppose $M$ is the number of spikes from these inactivated neurons and $V_{th}^{(\ell)}=1$. Then we can get
\begin{align}
	&\sum\limits _{t=0}^T s_{i}^{(\ell)}(t)=\operatorname{clip}\left(\sum_{t=0}^{T} I_{i}^{(\ell)}(t)-V_{i}^{(\ell)}(T), 0, T\right) \\
	&V_{i}^{(\ell)}(T)=\sum_{t=0}^{T} I_{i}^{(\ell)}(t)-\operatorname{clip}\left(\left\lfloor\sum\limits_{t=0}^{T} I^{(\ell)}_i(t)\right\rfloor, 0, T\right)-M
\end{align}
where $M \cdot \operatorname{clip}\left(\left\lfloor\sum\limits_{t=0}^{T} I^{(t)}_i(t)\right\rfloor, 0, T\right)=0$ is used to limit $M=0, if \quad a_i^{(\ell)}>0$. SIN arises from the fact that the sum of partial weights reaches the threshold in a short time, which can be reduced to some extent by increasing the parameter $p$. The problem is exacerbated during phase coding because more information is packed in one spike. Not that SIN doesn't matter very much in classification problems, because we only need to get a maximum for classification. The approximation degree of ANN activation value is required to be higher for detection or segmentation application.

In addition, most work uses average pooling, because it is equivalent to convolution operations with the same kernel size and step size. Some works with MaxPooling select spikes from neurons with maximum firing rates to pass through. This not only requires extra calculation of firing rate, but also causes \textbf{pooling errors} in the output as shown in the green box in Fig. \ref{error}. A key problem is that the neurons selected for this method are changed, namely, $\underset{\mathrm{i}}{\arg \max } \sum_{\tau=0}^{\mathrm{t}} \mathrm{s}_{\mathrm{i}}^{(\ell)}(\tau)$ is not a constant. Then we get 
\begin{align}
	\max (\mathbb{O})	\leq r_i^{(\ell)}\leq \operatorname{sum}(\mathbb{O})
	\label{o2o}
\end{align}
where $\mathbb{O}$ is the set of input values for the MaxPooling layer. Due to the instability of synaptic current, the neuron with the maximum firing rate may vary from neuron to neuron, resulting in the output tending to be larger than the actual activation value and smaller than the sum of the first two large values.

\begin{algorithm}[!b] 
	\caption{Transmit with Burst Spikes} 
	﻿\label{alg:algorithm}
	\begin{algorithmic}[1]
		\IF{$\operatorname{isinstance}$(layer, ReLU)}
		\STATE Calculate $s_i^{(\ell)}$ with Eq. (\ref{spike})
		\WHILE{ $out \leq \Gamma$ and $s_i^{(\ell)}=1$}
		\STATE Update $out \leftarrow out+s_i^{(\ell)}$
		\STATE  Update $V_i^{(\ell)}(t)\leftarrow V_i^{(\ell)}(t)-s_i^{(\ell)}$ using soft-reset
		\STATE Calculate $s_i^{(\ell)}$ with Eq. (\ref{spike})
		\ENDWHILE
		\ENDIF
	\end{algorithmic} 
\end{algorithm}

\subsection{Transmit with Burst Spikes}

\begin{figure}[t]
	\centering
	\includegraphics[scale=0.6]{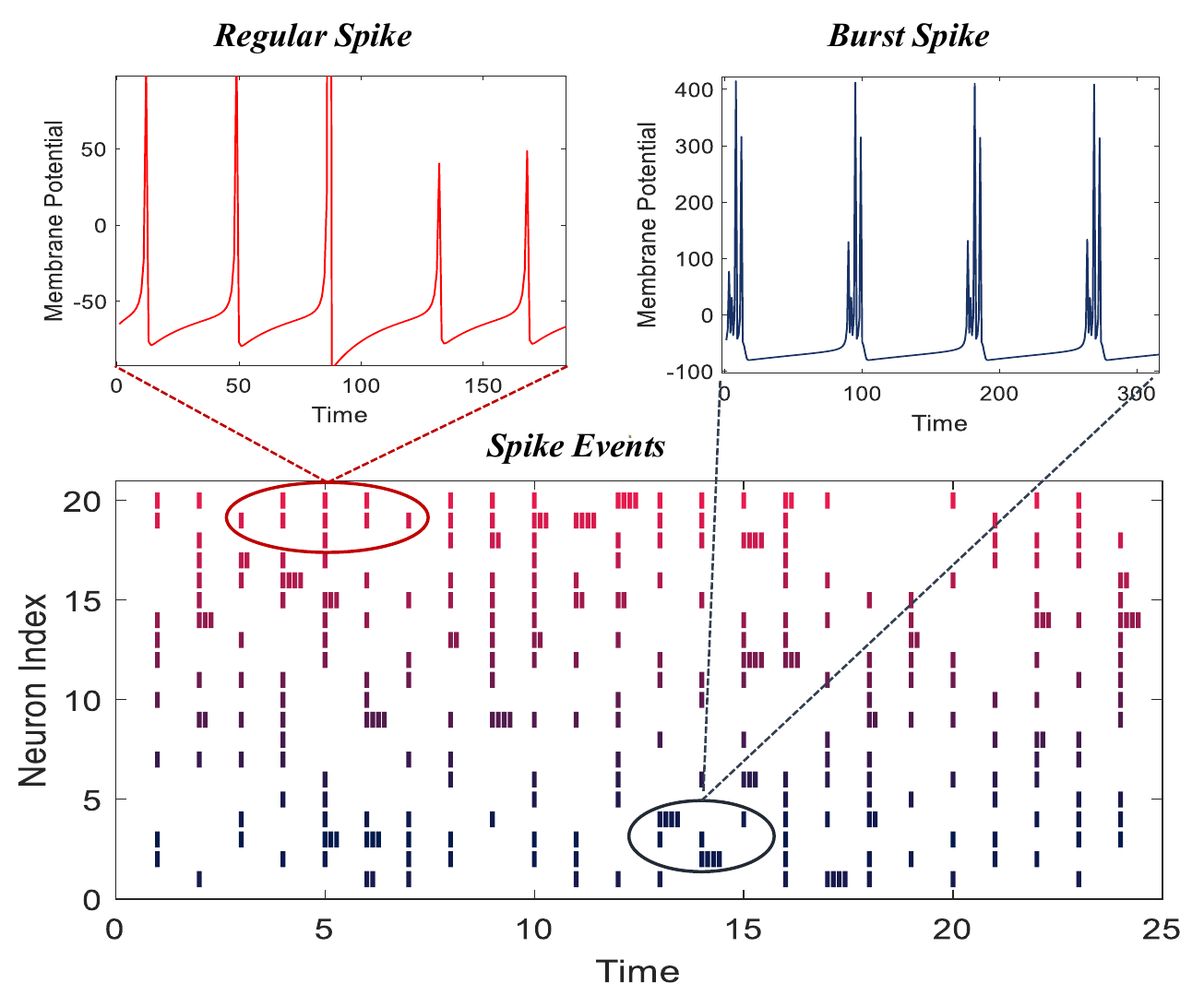}
	\caption{Illustration of burst spikes. Burst spikes can make the residual information in the membrane potential release between simulation steps, while the regular spikes can only release one spike at most at a limited simulation step.}
	\label{burst}
\end{figure}

In the conversion process, no matter the value of $p$, the converted SNN will be caught in a dilemma of speed and precision. As shown in Eq. (\ref{principle}), under the clock-driven simulation framework, the maximum firing rate $r^{(\ell)}$ is 1. However, in biological nervous systems and neuromorphic hardware, neurons process information asynchronously, updating between observable time steps. Burst neurons continuously emit spikes when stimulated by input current, increasing the reliability of information transmission. 

Inspired by the neuron burst mechanism, we introduce burst neurons into converted SNN. It can emit spikes between two simulation steps. Then the neuron of the next layer processes these spikes the next time step, as shown in Fig. \ref{burst}. Burst spikes can make the residual information in the membrane potential released between simulation steps. In contrast, the regular spikes can only release one spike at most at a limited simulation step.
Of course, considering the time needed for spike issuance in the actual situation, we limit the maximum burst spikes number $\Gamma$ to prevent interference to the next simulation step.
The pseudo-code is shown in Algorithm \ref{alg:algorithm}. 
The introduction of burst can make $\mathbb{R}_1$ set larger, that is, $\left\{j \mid 1 < \frac{o_{j}^{(\ell)}}{max_{p}^{(\ell)}} \leq \Gamma \right\}$ are saved from $\mathbb{R}_2$, while there is no change to $p$. 
Thus SNN can deliver the same amount of information at a faster way.

\subsection{LIPooling}
It can be seen from Eq. (\ref{o2o}) that the problem of using the spikes of the neuron with the maximum firing rate to carry out the MaxPooling is that the indexed neurons are changing, and there is no interaction between neurons. When multiple neurons have similar, even equal firing rates, previous work randomly selects the output of one neuron or uses the output of the neuron with the smallest index. These neurons will play the role of winners in different degrees for some time steps. Therefore, the output of the Maxpooling layer of converted SNN is usually greater than the actual maximum value. So most works in the past use Average pooling instead.

\begin{figure}[!t]
	\centering
	\includegraphics[scale=0.55]{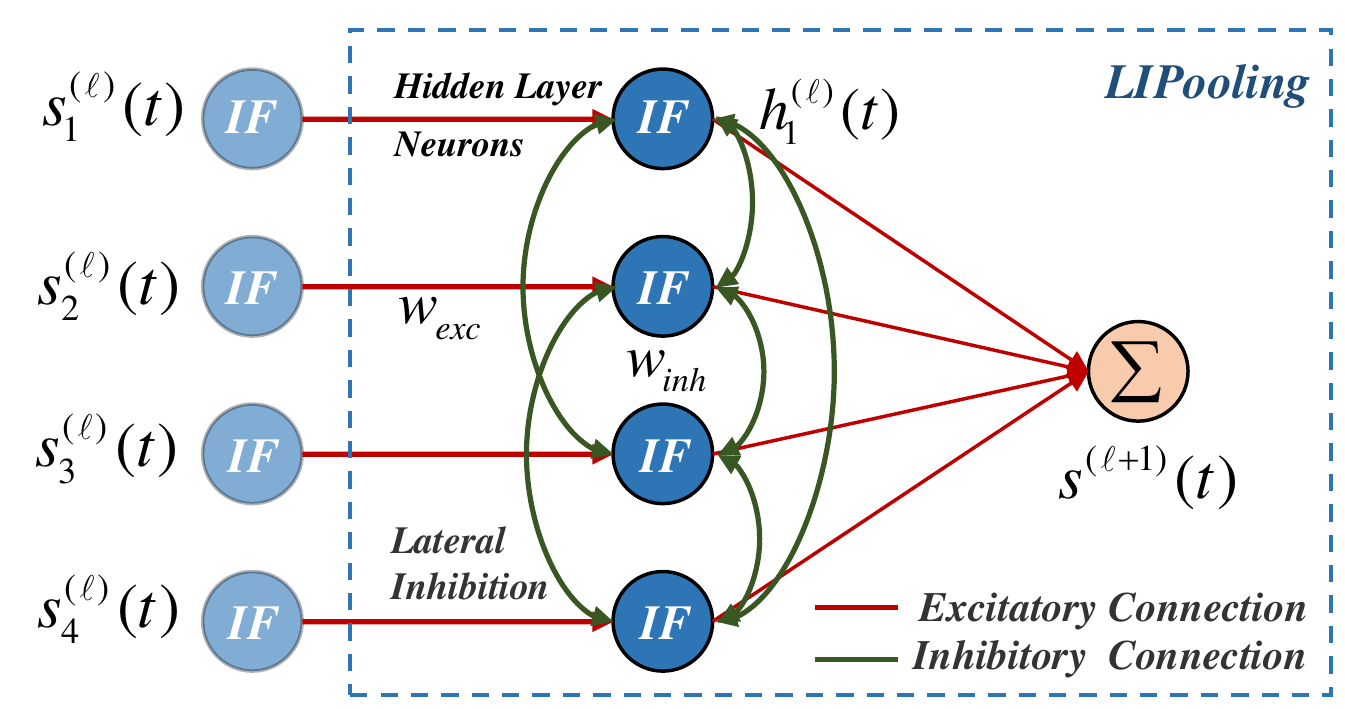}
	\caption{Illustration of LIPooling. Inhibition between hidden layer neurons implicitly selects neurons with the maximum firing rate and can correct historical output by suppressing the output spike.}
	\label{lipool}
\end{figure}

Inspired by the lateral inhibition mechanism \cite{koyama2018mutual} of biological neurons, we propose LIPooling method for converted SNN, as shown in Fig. \ref{lipool}. The number of hidden layer neurons of LIPooling is the same as the input of MaxPooling, and the updating of membrane potential and spike is shown in the Eq. (\ref{spike}) and (\ref{vm}). The excitatory connection receives the spikes $s_i^{(\ell)}(t)$ from the previous layer one to one, and the inhibitory connection influences the membrane potential of other neurons. Then the output neurons sum up the spikes of the hidden layer neurons. In fact, due to historical inhibition, only one neuron spikes at most at any time step, that is, the neuron with the maximum firing rate is selected to pass through, and its membrane potential update can be described as follows:
\begin{align}
	V_{hi}^{(\ell)}(t) = V_{hi}^{(\ell)}(t-1) + w_{exc}s_i^{(\ell-1)}(t) + \textbf{w}_{inh}\textbf{h}^{(\ell)}(t-1)
\end{align}
where $V_{hi}^{(\ell)}$ and $\textbf{h}^{(\ell)}$ represent potential and spikes of hidden layers respectively, $w_{exc}$ and $\textbf{w}_{inh}$ represent excitatory and inhibitory connections respectively. Here we set $w_{exc}=1, \textbf{w}_{inh}=-1, V_{th}=1$. Then we get the output $s_i^{(\ell+1)}(t) = \sum \limits _j h_j^{(\ell)}(t)$. It should be noted that, by adding the lateral inhibition mechanism, the spikes emitted by neurons with the maximum firing rate are suppressed by LIPooling to offset the spikes emitted by other neurons in the previous simulation time. Then we can make $\sum\limits_{t=0}^T s^{(\ell+1)}(t)=\max\limits_{i}\{\sum\limits_{t=0}^T s_i^{(\ell)}(t)\}$.

\section{Experiments}

\subsection{Experimental Setup}

To demonstrate the effectiveness and efficiency of our algorithm, we conduct experiments on CIFAR and ImageNet. For CIFAR, we use stochastic gradient descent with 0.9 momentum for weight optimization. The cosine learning rate decay strategy with an initial value of 0.1 is used to change the learning rate dynamically. The network is optimized for 300 iterations with a batch size of 128. We use data augmentations for high performance. For ImageNet, we use a pre-trained model from pytorchcv\footnote{\href{https://pypi.org/project/pytorchcv/}{https://pypi.org/project/pytorchcv/}}. We follow the previous work and use the same network structure to ensure a fair comparison.  For CIFAR, we use VGG16 and ResNet20 models, and for ImageNet, we use VGG16 model. As section \ref{32} described, SIN has less effect on the classification problem. Future work will explore the influence of the SIN problem on detection tasks.

\begin{figure}
	\centering
	\begin{minipage}{0.99\linewidth}
		\subfigure[Accuracy curve compared to the baseline.]{
			\label{a}
			\includegraphics[scale=0.57]{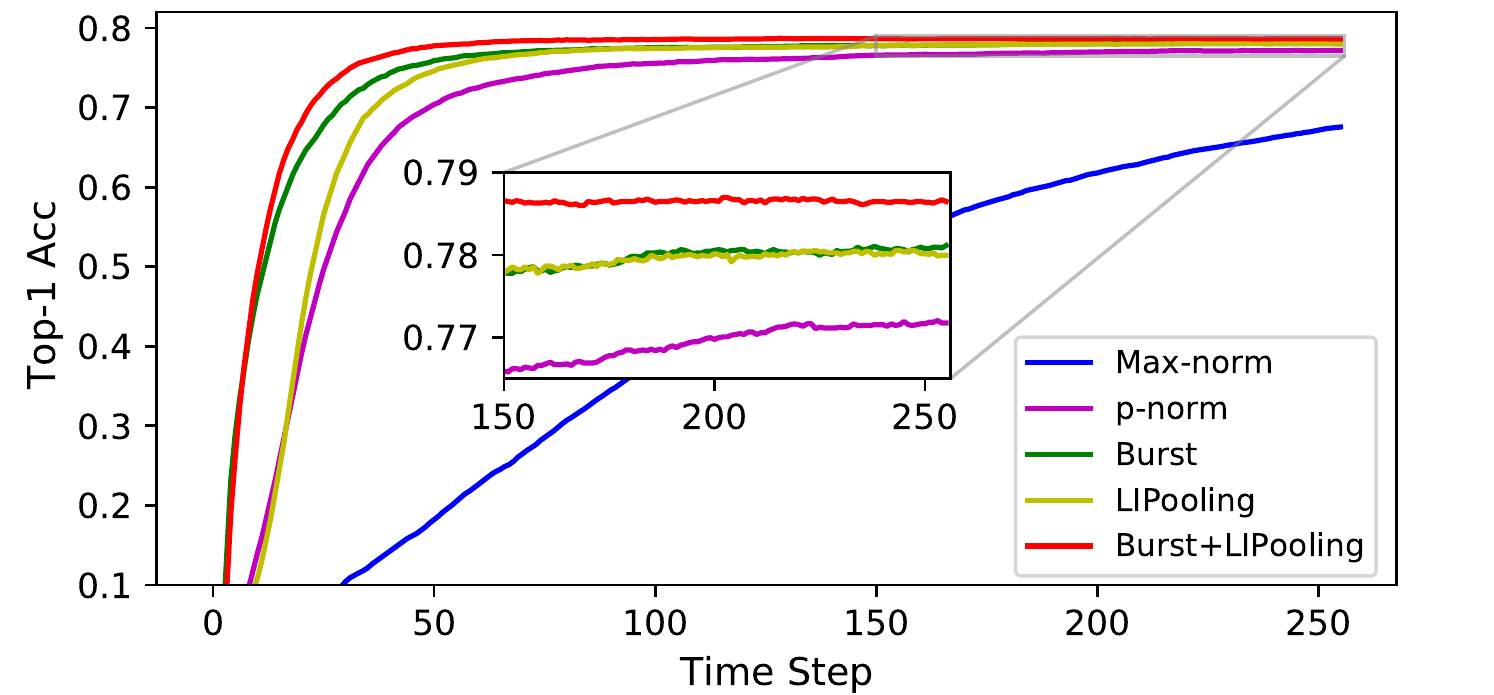}}
	\end{minipage}\\
	\begin{minipage}{1\linewidth}
		\subfigure[Correct ratio.]{
			\label{b}
			\includegraphics[scale=0.28]{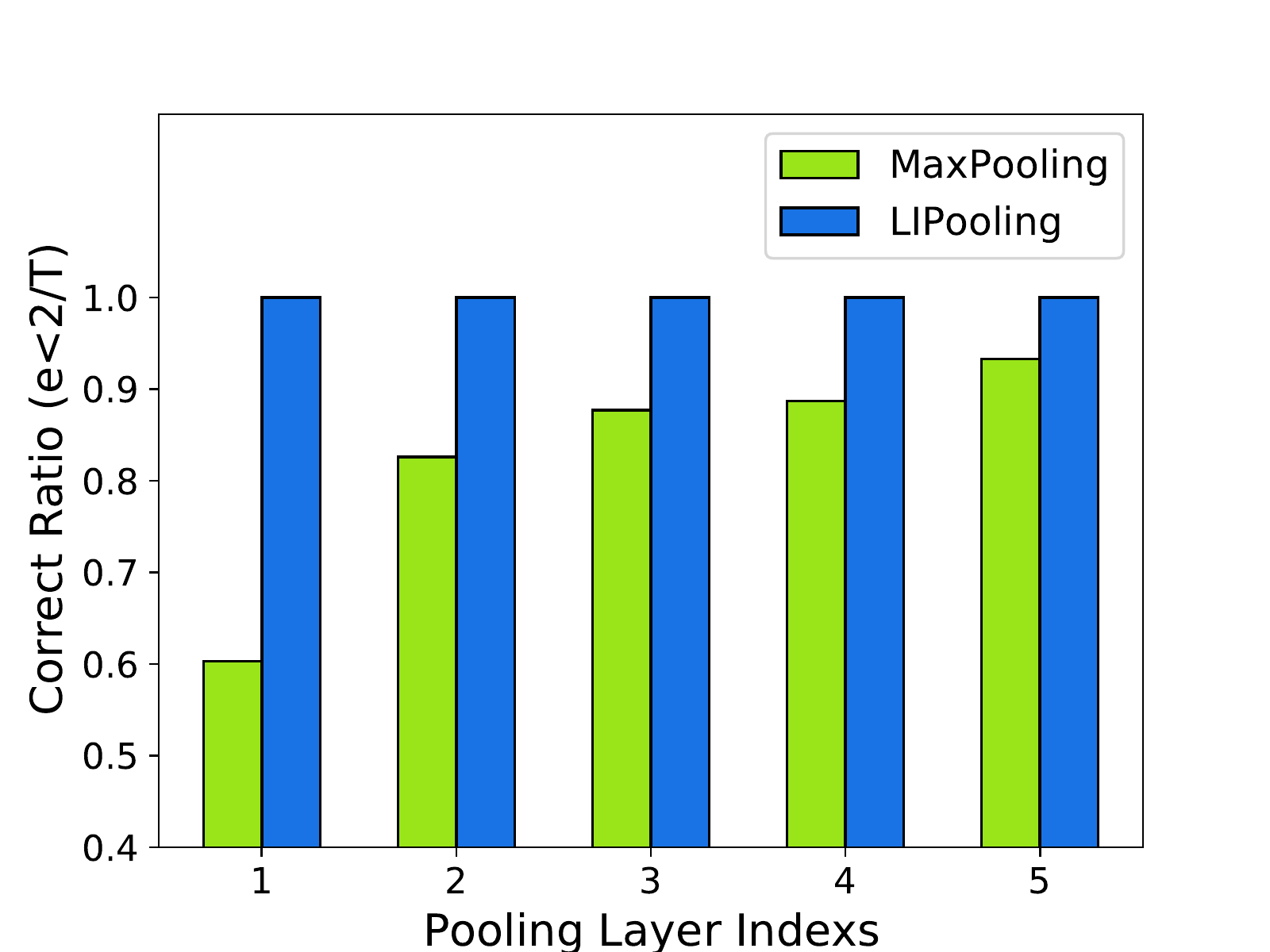}}
		\subfigure[Firing rate in SNN.]{
			\label{c}
			\includegraphics[scale=0.28]{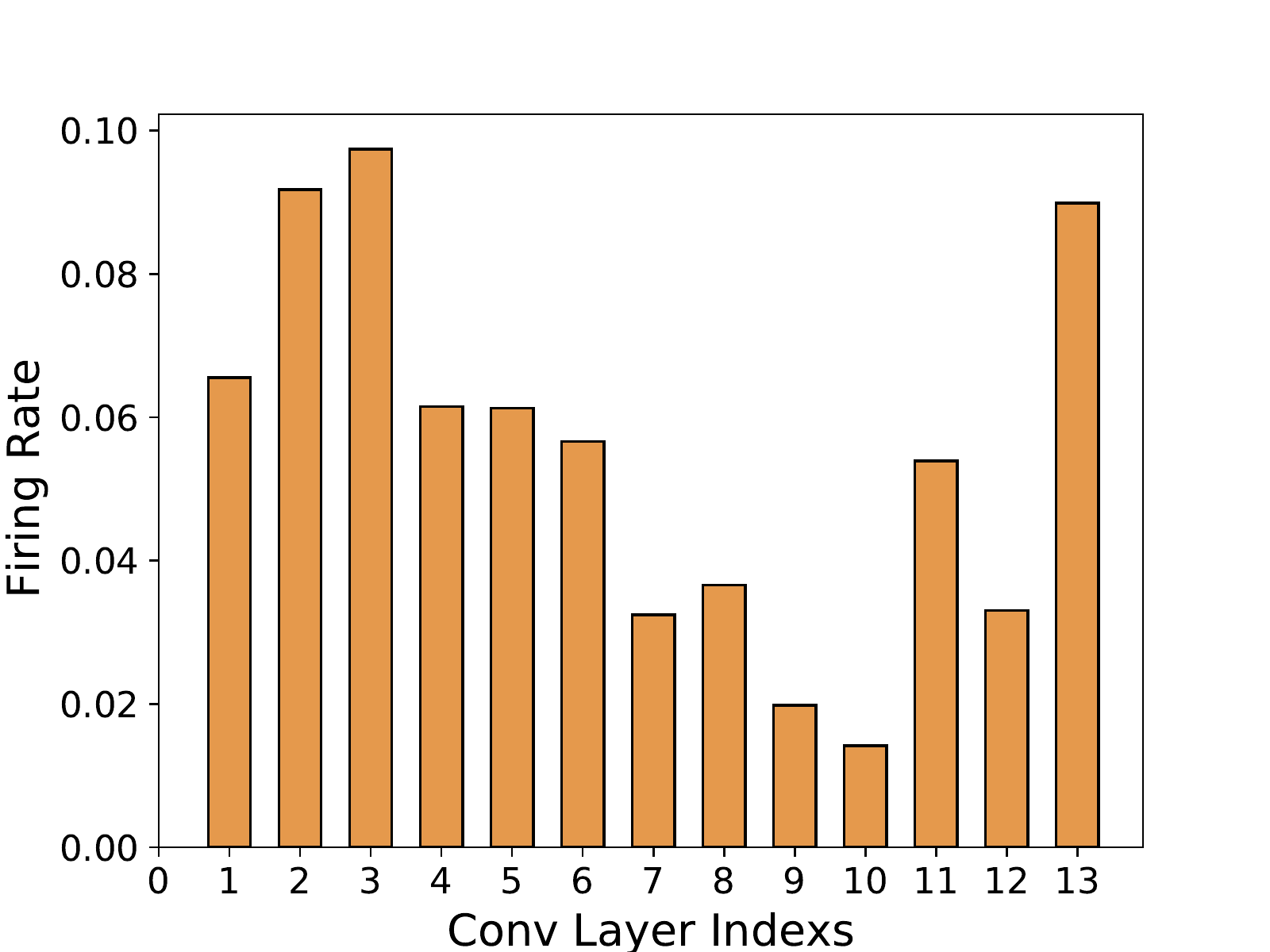}}
	\end{minipage}
	\caption{Experiments on CIFAR100 with VGG16. (a) demonstrates that our algorithm can significantly improve the accuracy and speed. (b) indicates that the lateral inhibition effect of LIPooling can solve the errors introduced by MaxPooling in SNN. (c) visualizes the firing rate of the Conv layers}
	\label{ablation}
\end{figure}

\begin{table}[!t]
	\centering
		\begin{tabular}{lcccc}
			\toprule
			\multirow{2}{*}{\textbf{Method}} & \multicolumn{2}{c}{\textbf{VGG16 (78.49)} } &  \multicolumn{2}{c}{\textbf{ResNet20 (80.69)} }\\ \cmidrule(r){2-3} \cmidrule(r){4-5}
			&$T=16$&$T=32$&$T=16$&$T=32$\\ 
			\midrule
			Max-norm 		   & 2.99   &  10.91 &  3.16 & 27.78 \\
			p-Norm    			  & 25.72 &  58.51 &  8.82 & 45.83  \\
			\textbf{LIPooling} 	   & 24.37  &  \textbf{65.63} &  \textbf{12.73} & \textbf{53.35} \\ 
			\textbf{Burst ($\Gamma=2$)} 	& 54.88   &  70.96 & 38.13 & 67.85 \\
			\textbf{Burst ($\Gamma=5$)} 	& \textbf{57.21}   &  \textbf{71.41} & \textbf{58.1}0 & \textbf{76.30} \\
			\bottomrule
		\end{tabular}
		\caption{Experiments on CIFAR100 with VGG16 and ResNet20.}
		\label{tab1}
	\end{table}

	\subsection{Ablation Study}
	
	\begin{table*}[!t]
		\centering
		\resizebox{\textwidth}{!}{
			\begin{tabular}{lllcccccc}
				\toprule
				\textbf{Dataset} & \textbf{Method} & \textbf{Network}   &  \textbf{ANN}  & \textbf{SNN Best}& $T=32$ & $T=64$& $T=128$ & $T=256$ \\
				\midrule
				\multirow{6}{*}{\textbf{CIFAR10}}
				& SpikeNorm \cite{sengupta2019going} & VGG16 & 91.70&91.55& -& -& -& -\\
				& Hybrid \cite{rathi2020enabling}& VGG16 &92.81& 92.48 & -& -& 91.13 &- \\
				& RMP-SNN \cite{han2020rmp} & VGG16 & 93.63 & 93.63 &60.30 &90.35  &92.41 &93.04\\
				& \textbf{This Work (Burst+LIPooling)} & \textbf{VGG16} & \textbf{95.74} & \textbf{95.75} & \textbf{95.58} & \textbf{95.66} & \textbf{95.69} & \textbf{95.72}\\
				& RMP-SNN \cite{han2020rmp} & ResNet20 & 91.47& 91.36 &  -& - &87.60 &89.37\\
				& \textbf{This Work (Burst+LIPooling)} & \textbf{ResNet20} & \textbf{96.56} & \textbf{96.59} & \textbf{96.11} & \textbf{96.49} & \textbf{96.45} & \textbf{96.36}\\
				
				\midrule
				\multirow{10}{*}{\textbf{CIFAR100}}
				& TSC \cite{han2020deep}          & VGG16 & 71.22 & 70.97 & - & - & 69.86 & 70.65\\ 
				& RMP-SNN \cite{han2020rmp} & VGG16 & 71.22 & 70.93 & - & - & 63.76 & 68.34\\
				& Opt. \cite{deng2021optimal}    & VGG16 & 77.89 & 77.71& 7.64   & 21.84 &55.04 &73.54\\
				& Calibration \cite{li2021free}    & VGG16  & 77.89 &77.87 & 73.55 &76.64 & 77.40   &77.68\\
				& \textbf{This Work (Burst+LIPooling)} & \textbf{VGG16} & \textbf{78.49} & \textbf{78.71} & \textbf{74.98} & \textbf{78.26} & \textbf{78.66} & \textbf{78.65}\\
				& TSC \cite{han2020deep}          & ResNet20 & 68.72 & 68.18 & - & - & 58.42 & 65.27\\ 
				& RMP-SNN \cite{han2020rmp} & ResNet20 & 68.72 & 67.82 & 27.64 & 46.91 & 57.69 & 64.06 \\
				& Opt. \cite{deng2021optimal} & ResNet20 & 77.16 & 77.22 & 51.27 & 70.12 & 75.81 & 77.22 \\
				& Calibration \cite{li2021free}    & ResNet20  & 77.16 &77.73 & 76.32  &77.29  &77.73 &77.63\\
				& \textbf{This Work (Burst+LIPooling)} & \textbf{ResNet20} & \textbf{80.69} & \textbf{80.72} & \textbf{76.39} & \textbf{79.83} & \textbf{80.52} & \textbf{80.57}\\
				
				\midrule
				\multirow{6}{*}{\textbf{ImageNet}}
				& Hybrid \cite{rathi2020enabling} & VGG16 & 69.35 & 65.19 &- &- &-&62.73 \\
				& TSC \cite{han2020deep}          & VGG16 & 73.49 & 73.46 & - & - & - & 69.71\\ 
				& RMP-SNN \cite{han2020rmp} & VGG16 & 73.49  & 73.09&- & - & - & 48.32\\
				& Opt. \cite{deng2021optimal}    & VGG16 & 75.36 &73.88 & 0.114 &0.118 &0.122 &1.81\\
				& Calibration \cite{li2021free}    & VGG16  & 75.36 & 75.32 &63.64 & 70.69 & 73.32 &74.23\\
				& \textbf{This Work (Burst+LIPooling)} & \textbf{VGG16} & \textbf{74.27} & \textbf{74.26} & \textbf{70.61} & \textbf{73.32} & \textbf{73.99} & \textbf{74.25}\\
				\bottomrule
		\end{tabular}}
		\caption{Classification accuracy on CIFAR and ImageNet for our converted SNNs, and compared to other conversion methods and ANN.}
		\label{tab3}
	\end{table*}
	
	To verify the effectiveness of the proposed method, we perform ablation experiments on CIFAR100 with VGG16 and ResNet20. All nets are trained with BN layers. 
	
	We first verify the effect of burst spikes on conversion. In Tab. \ref{tab1}, our method has the lowest reasoning time and achieves higher performance compared to the regular spike based method. For example, when converting VGG16, our method requires only 32 time steps to get 71.41\% accuracy. The network has different performances under different $\Gamma$, and the number of $\Gamma$ is limited by the processing capacity of the neuromorphic hardware. The accuracy curve of the whole conversion process is shown in Fig. \ref{a}. Burst significantly improves the accuracy and speed of conversion compared to baseline.
	
	We also verify the influence of LIPooling on conversion. To make the comparison fair, p-Norm with MaxPooling is used. LIPooling perfectly solves the problem of MaxPooling in SNN, thus significantly improving the accuracy and speed as shown in Tab. \ref{tab1}. We calculate the error between the output firing rate and the value that directly pool the firing rate of  input layer, and count the ratio of positive activation values correctly issued when the allowable error $e<\frac{2}{T}$. Fig. \ref{b} demonstrates that LIPooling can accurately pass the maximum value according to the spikes after LIPooling.

	\subsection{Comparison with Other Works}
	
	In this section, we compare our proposed method with other existing SOTA conversion works. Our work shows \textbf{higher and faster} performance, as shown in the Tab. \ref{tab3}.
	First, we focus on the SNN performance of the conversion. For VGG16 and ResNet, our conversion method achieves a performance loss of less than 1\% over 100 time steps. For CIFAR10 and CIFAR100, we achieve 96.59\% (ResNet20) and 80.72\% (ResNet20) performance, respectively, which is the best performance of SNN on these datasets.
For ImageNet, the accuracy of 74.26\% has obvious advantages over RMP \cite{han2020rmp} and other methods. It is also close to the SNN calibration \cite{li2021free} method.
In addition to the superior performance, our method has outstanding advantages in simulation time. For CIFAR10, we achieve 96.11\% performance with only 32 time steps. We also need dozens of time steps to achieve acceptable performance (78.26\% on VGG16 with $T=64$) on CIFAR100. For ImageNet, our method is twice as fast as the SNN calibration method, achieving a performance of 73.99\% with 128 time steps.

	\subsection{Energy Efficiency}
	In SNN, neurons that do not emit spikes do not participate in calculations, so the implementation of SNN on hardware has the characteristics of energy efficiency, especially on neuromorphic hardware. We select VGG16 to verify the energy efficiency of the proposed method on CIFAR100 and use the energy estimation method in \cite{rathi2021diet}. Generally, a 32-bit floating-point MAC operation consumes 4.6J, while the AC operation only consumes	0.9J. We calculate the mean firing rate of each layer on the whole validation set. Although burst spikes are used to improve the firing capacity of neurons, we can see from Fig. \ref{c} that the maximum and minimum firing rates of each layer are 0.097 and 0.014, respectively, which reflects the sparsity of spikes activity.
	Our spiking VGG16 only costs 69.30\% energy of ANN's consumption while getting SOTA performance with extra short time,  which exhibits the properties of low time latency, high inference performance, and energy efficiency.

	\section{Conclusion}

	In this paper, we firstly theoretically analyze the nodes that may produce errors in conversion. Furthermore, we propose the neuron model for releasing burst spikes to prevent information residue and replacing the MaxPooling with LIPooling to ensure the accurate transmission of information. To the best of our knowledge, this is the first time to analyze the error from the whole conversion process, including the differences between ReLU and IF, time dimension, and pooling operation. Our method can achieve comparable accuracy with ANN in a short ($<100$) simulation time. It can successfully convert VGGG16 on ImageNet, and our method also achieves the best performance on networks using BN and MaxPooling. Hopefully, our work can be combined with neuromorphic hardware to demonstrate the advantages of SNNs. 
	
	\newpage
	\section*{Acknowledgements}
	
	This study is supported by National Key Research and Development Program (2020AAA0107800), the Strategic Priority Research Program of the Chinese Academy of Sciences (Grant No. XDB32070100).
	
	\bibliographystyle{named}
	\bibliography{ijcai22}

\begin{thebibliography}{}

\bibitem[\protect\citeauthoryear{Akopyan \bgroup \em et al.\egroup
  }{2015}]{akopyan2015truenorth}
Filipp Akopyan, Jun Sawada, Andrew Cassidy, Rodrigo Alvarez-Icaza, John Arthur,
  Paul Merolla, Nabil Imam, Yutaka Nakamura, Pallab Datta, Gi-Joon Nam, et~al.
\newblock Truenorth: Design and tool flow of a 65 mw 1 million neuron
  programmable neurosynaptic chip.
\newblock {\em IEEE TCAD}, 34(10):1537--1557, 2015.

\bibitem[\protect\citeauthoryear{Cao \bgroup \em et al.\egroup
  }{2015}]{cao2015spiking}
Yongqiang Cao, Yang Chen, and Deepak Khosla.
\newblock Spiking deep convolutional neural networks for energy-efficient
  object recognition.
\newblock {\em International Journal of Computer Vision}, 113(1):54--66, 2015.

\bibitem[\protect\citeauthoryear{Davies \bgroup \em et al.\egroup
  }{2018}]{davies2018loihi}
Mike Davies, Narayan Srinivasa, Tsung-Han Lin, Gautham Chinya, Yongqiang Cao,
  Sri~Harsha Choday, Georgios Dimou, Prasad Joshi, Nabil Imam, Shweta Jain,
  et~al.
\newblock Loihi: A neuromorphic manycore processor with on-chip learning.
\newblock {\em IEEE Micro}, 38(1):82--99, 2018.

\bibitem[\protect\citeauthoryear{Deng and Gu}{2021}]{deng2021optimal}
Shikuang Deng and Shi Gu.
\newblock Optimal conversion of conventional artificial neural networks to
  spiking neural networks.
\newblock {\em arXiv preprint arXiv:2103.00476}, 2021.

\bibitem[\protect\citeauthoryear{Diehl \bgroup \em et al.\egroup
  }{2015}]{diehl2015fast}
Peter~U Diehl, Daniel Neil, Jonathan Binas, Matthew Cook, Shih-Chii Liu, and
  Michael Pfeiffer.
\newblock Fast-classifying, high-accuracy spiking deep networks through weight
  and threshold balancing.
\newblock In {\em IJCNN}, pages 1--8. ieee, 2015.

\bibitem[\protect\citeauthoryear{Ding \bgroup \em et al.\egroup
  }{2021}]{ding2021optimal}
Jianhao Ding, Zhaofei Yu, Yonghong Tian, and Tiejun Huang.
\newblock Optimal ann-snn conversion for fast and accurate inference in deep
  spiking neural networks.
\newblock {\em arXiv preprint arXiv:2105.11654}, 2021.

\bibitem[\protect\citeauthoryear{Han and Roy}{2020}]{han2020deep}
Bing Han and Kaushik Roy.
\newblock Deep spiking neural network: Energy efficiency through time based
  coding.
\newblock In {\em ECCV}, pages 388--404. Springer, 2020.

\bibitem[\protect\citeauthoryear{Han \bgroup \em et al.\egroup
  }{2020}]{han2020rmp}
Bing Han, Gopalakrishnan Srinivasan, and Kaushik Roy.
\newblock Rmp-snn: Residual membrane potential neuron for enabling deeper
  high-accuracy and low-latency spiking neural network.
\newblock In {\em CVPR}, pages 13558--13567, 2020.

\bibitem[\protect\citeauthoryear{Kim \bgroup \em et al.\egroup
  }{2018}]{kim2018deep}
Jaehyun Kim, Heesu Kim, Subin Huh, Jinho Lee, and Kiyoung Choi.
\newblock Deep neural networks with weighted spikes.
\newblock {\em Neurocomputing}, 311:373--386, 2018.

\bibitem[\protect\citeauthoryear{Kim \bgroup \em et al.\egroup
  }{2020}]{kim2020spiking}
Seijoon Kim, Seongsik Park, Byunggook Na, and Sungroh Yoon.
\newblock Spiking-yolo: Spiking neural network for energy-efficient object
  detection.
\newblock In {\em AAAI}, volume~34, pages 11270--11277, 2020.

\bibitem[\protect\citeauthoryear{Koyama and Pujala}{2018}]{koyama2018mutual}
Minoru Koyama and Avinash Pujala.
\newblock Mutual inhibition of lateral inhibition: a network motif for an
  elementary computation in the brain.
\newblock {\em Current opinion in neurobiology}, 49:69--74, 2018.

\bibitem[\protect\citeauthoryear{Li \bgroup \em et al.\egroup
  }{2021a}]{li2021bsnn}
Yang Li, Yi~Zeng, and Dongcheng Zhao.
\newblock Bsnn: Towards faster and better conversion of artificial neural
  networks to spiking neural networks with bistable neurons.
\newblock {\em arXiv preprint arXiv:2105.12917}, 2021.

\bibitem[\protect\citeauthoryear{Li \bgroup \em et al.\egroup
  }{2021b}]{li2021free}
Yuhang Li, Shikuang Deng, Xin Dong, Ruihao Gong, and Shi Gu.
\newblock A free lunch from ann: Towards efficient, accurate spiking neural
  networks calibration.
\newblock {\em arXiv preprint arXiv:2106.06984}, 2021.

\bibitem[\protect\citeauthoryear{Luo \bgroup \em et al.\egroup
  }{2020}]{luo2020siamsnn}
Yihao Luo, Min Xu, Caihong Yuan, Xiang Cao, Liangqi Zhang, Yan Xu, Tianjiang
  Wang, and Qi~Feng.
\newblock Siamsnn: Siamese spiking neural networks for energy-efficient object
  tracking.
\newblock {\em arXiv preprint arXiv:2003.07584}, 2020.

\bibitem[\protect\citeauthoryear{Maass}{1997}]{maass1997networks}
Wolfgang Maass.
\newblock Networks of spiking neurons: the third generation of neural network
  models.
\newblock {\em Neural networks}, 10(9):1659--1671, 1997.

\bibitem[\protect\citeauthoryear{Rathi and Roy}{2021}]{rathi2021diet}
Nitin Rathi and Kaushik Roy.
\newblock Diet-snn: A low-latency spiking neural network with direct input
  encoding and leakage and threshold optimization.
\newblock {\em IEEE TNNLS}, 2021.

\bibitem[\protect\citeauthoryear{Rathi \bgroup \em et al.\egroup
  }{2020}]{rathi2020enabling}
Nitin Rathi, Gopalakrishnan Srinivasan, Priyadarshini Panda, and Kaushik Roy.
\newblock Enabling deep spiking neural networks with hybrid conversion and
  spike timing dependent backpropagation.
\newblock {\em arXiv preprint arXiv:2005.01807}, 2020.

\bibitem[\protect\citeauthoryear{Roy \bgroup \em et al.\egroup
  }{2019}]{roy2019towards}
Kaushik Roy, Akhilesh Jaiswal, and Priyadarshini Panda.
\newblock Towards spike-based machine intelligence with neuromorphic computing.
\newblock {\em Nature}, 575(7784):607--617, 2019.

\bibitem[\protect\citeauthoryear{Rueckauer \bgroup \em et al.\egroup
  }{2017}]{rueckauer2017conversion}
Bodo Rueckauer, Iulia-Alexandra Lungu, Yuhuang Hu, Michael Pfeiffer, and
  Shih-Chii Liu.
\newblock Conversion of continuous-valued deep networks to efficient
  event-driven networks for image classification.
\newblock {\em Frontiers in neuroscience}, 11:682, 2017.

\bibitem[\protect\citeauthoryear{Sengupta \bgroup \em et al.\egroup
  }{2019}]{sengupta2019going}
Abhronil Sengupta, Yuting Ye, Robert Wang, Chiao Liu, and Kaushik Roy.
\newblock Going deeper in spiking neural networks: Vgg and residual
  architectures.
\newblock {\em Frontiers in neuroscience}, 13:95, 2019.

\bibitem[\protect\citeauthoryear{Tan \bgroup \em et al.\egroup
  }{2020}]{tan2020strategy}
Weihao Tan, Devdhar Patel, and Robert Kozma.
\newblock Strategy and benchmark for converting deep q-networks to event-driven
  spiking neural networks.
\newblock {\em arXiv preprint arXiv:2009.14456}, 2020.

\bibitem[\protect\citeauthoryear{Wu \bgroup \em et al.\egroup
  }{2018}]{wu2018spatio}
Yujie Wu, Lei Deng, Guoqi Li, Jun Zhu, and Luping Shi.
\newblock Spatio-temporal backpropagation for training high-performance spiking
  neural networks.
\newblock {\em Frontiers in neuroscience}, 12:331, 2018.

\bibitem[\protect\citeauthoryear{Zeng \bgroup \em et al.\egroup
  }{2017}]{zeng2017improving}
Yi~Zeng, Tielin Zhang, and Bo~Xu.
\newblock Improving multi-layer spiking neural networks by incorporating
  brain-inspired rules.
\newblock {\em Science China Information Sciences}, 60(5):1--11, 2017.

\bibitem[\protect\citeauthoryear{Zhang \bgroup \em et al.\egroup
  }{2019}]{zhang2019tdsnn}
Lei Zhang, Shengyuan Zhou, Tian Zhi, Zidong Du, and Yunji Chen.
\newblock Tdsnn: From deep neural networks to deep spike neural networks with
  temporal-coding.
\newblock In {\em AAAI}, volume~33, pages 1319--1326, 2019.

\end{thebibliography}

\end{document}